\documentclass[12pt,a4paper]{article}
\usepackage[english]{babel}

\usepackage{amsfonts}
\usepackage{authblk}
\usepackage{fullpage}

\usepackage{amsmath}
\usepackage{amssymb}
\usepackage{soul}

\usepackage[linesnumbered,ruled]{algorithm2e}

\newcount\Comments
\usepackage{color}
\definecolor{darkgreen}{rgb}{0,0.5,0}
\definecolor{purple}{rgb}{1,0,1}
\newcommand{\kibitz}[2]{\ifnum\Comments=1\textcolor{#1}{#2}\fi}

\newcommand{\ie}{i.\,e.,\ }
\newcommand{\comment}[1]{}

\usepackage{graphicx}

\usepackage{subfig}

\usepackage{wrapfig}

\usepackage{acronym}
\acrodef{OCT}{Optical Coherence Tomography}
\acrodef{NFL}{Nerve Fibre Layer}
\acrodef{GCL}{Ganglion Cell Layer}
\acrodef{IPL}{Inner Plexiform Layer}
\acrodef{OPL}{Outer Plexiform Layer}
\acrodef{ONL}{Outer Nuclear Layer}
\acrodef{INL}{Inner Nuclear Layer}
\acrodef{RPE}{Retinal Pigment Epithelium}
\acrodef{ILM}{Internal Limiting Membrane}
\acrodef{MSE}{Mean Square Error}
\acrodef{EDI}{Enhanced Depth Imaging}
\acrodef{CNN}{Convolutional Neural Network}
\acrodef{CNNs}{Convolutional Neural Networks}
\acrodef{AMD}{Age-related Macular Degeneration}
\acrodef{DR}{Diabetic Retinopathy}
\acrodef{CSC}{Central Serous Chorioretinopathy}
\acrodef{BM}{Bruch's Membrane}
\acrodef{SNR}{Signal to Noise Ratio}
\acrodef{ReLU}{Rectified Linear Unit}
\acrodef{IS/OS}{Inner Segment/Outer Segment}
\acrodef{MRF}{Markov Random Field}

\usepackage[separate-uncertainty=true, multi-part-units=single, list-units=single]{siunitx}
\DeclareSIUnit\fps{Hz}
\DeclareSIUnit\px{pixels}
\DeclareSIUnit\bpp{bits/pixel}
\DeclareSIUnit\mb{MB}
\DeclareSIUnit\gb{GB}
\DeclareSIUnit\ghz{GHz}
\DeclareSIUnit\micrometer{\mu m}

\title{Pathological OCT Retinal Layer Segmentation using Branch Residual U-shape Networks}
\author[1]{Stefanos Apostolopoulos}
\author[2]{Sandro De Zanet}
\author[2]{Carlos Ciller}
\author[3]{Sebastian Wolf}
\author[1]{Raphael Sznitman}

\affil[1]{University of Bern, Switzerland}
\affil[2]{RetinAI Medical GmbH, Switzerland}
\affil[3]{Bern University Hospital, Switzerland}

\date{}

\begin{document}

\maketitle
\begin{abstract}
The automatic segmentation of retinal layer structures enables clinically-relevant quantification and monitoring of eye disorders over time in OCT imaging. Eyes with late-stage diseases are particularly challenging to segment, as their shape is highly warped due to pathological biomarkers. In this context, we propose a novel fully-\ac{CNN} architecture which combines dilated residual blocks in an asymmetric U-shape configuration, and can segment multiple layers of highly pathological eyes in one shot. We validate our approach on a dataset of late-stage AMD patients and demonstrate lower computational costs and higher performance compared to other state-of-the-art methods.
\end{abstract}

\section{Introduction}
\begin{figure}
\centering
\includegraphics[width=0.43\textwidth]{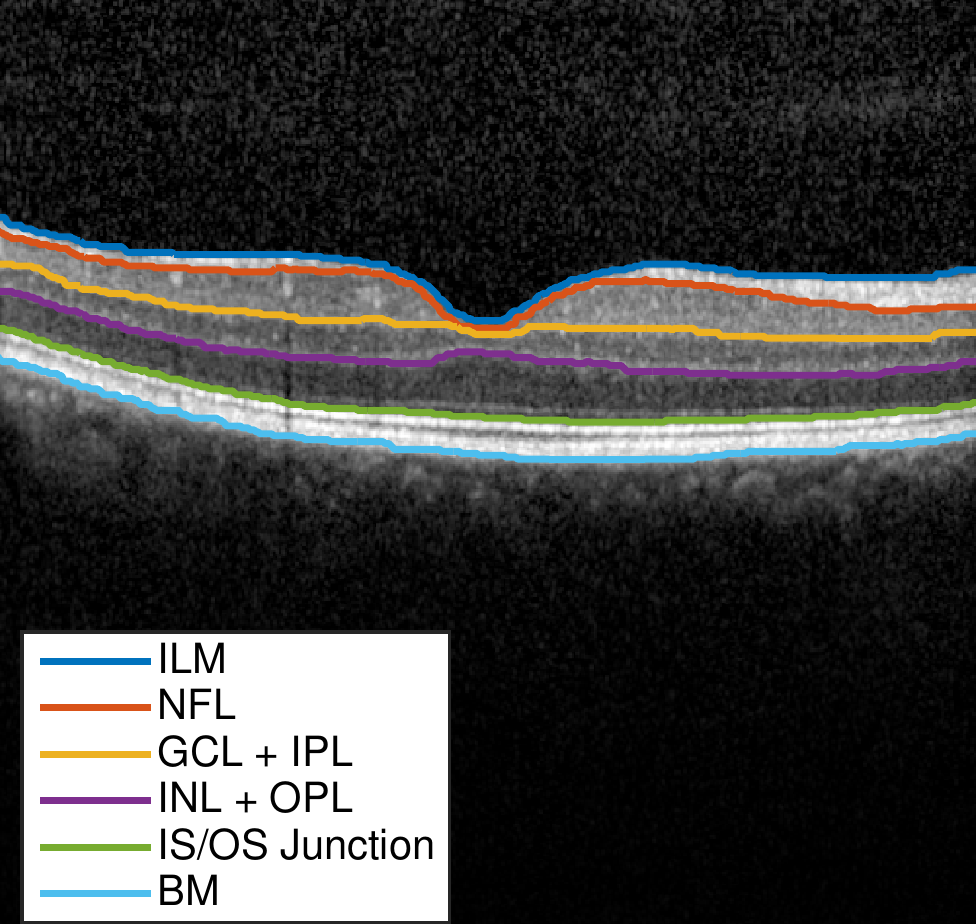}
\includegraphics[width=0.43\textwidth]{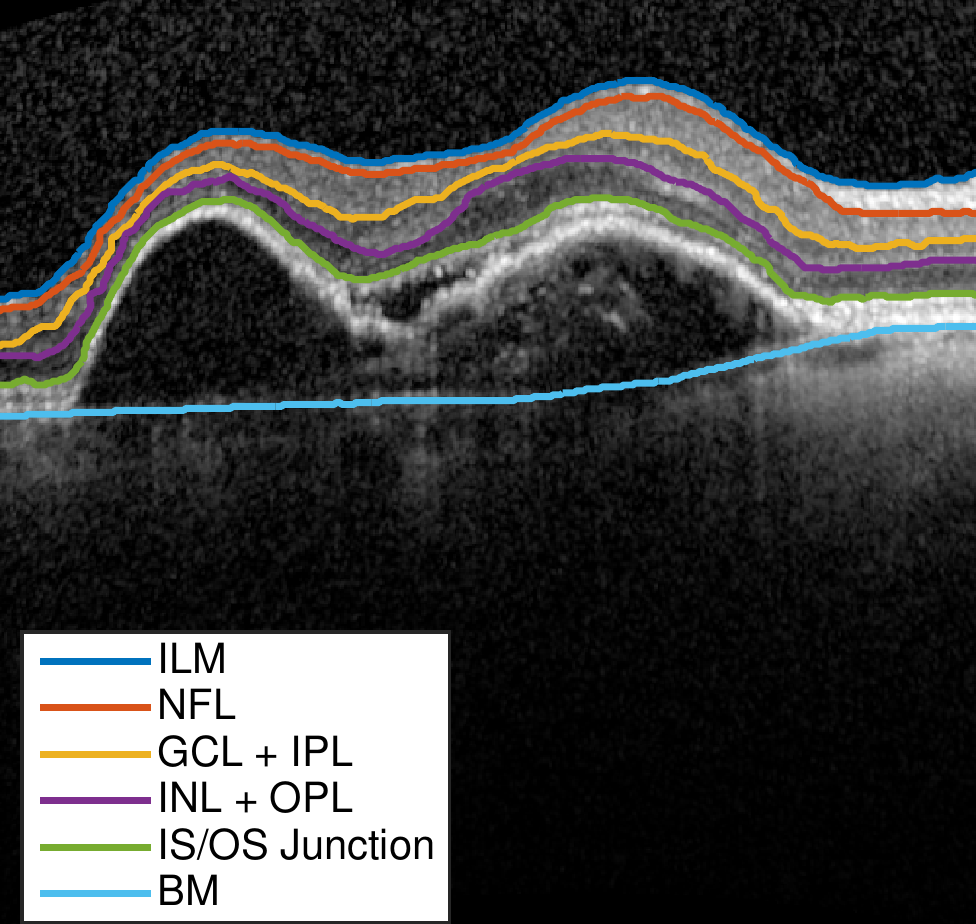}
\caption{Example of OCT cross-sections with retinal layer boundaries highlighted for (left) healthy subject and (right) late-stage AMD patient. The images were manually annotated by an expert ophthalmologist.}
\label{fig:examples}
\end{figure}

\ac{OCT} is a non-invasive medical imaging modality that provides micrometer-resolution volumetric scans of biological tissue~\cite{Huang1991}. Since its introduction in 1991,~\ac{OCT} has seen widespread use in the field of ophthalmology, as it enables direct, non-invasive imaging of the retinal layers. As shown in Fig.~\ref{fig:examples},~\ac{OCT} allows for the visualization of both healthy tissue and pathological biomarkers such as drusen, cysts and fluid pockets within and underneath the retinal layers. Critically, these have been linked to diseases such as \ac{AMD}, \ac{DR} and \ac{CSC}~\cite{Abramoff2010,Morgan2016}.

Given the widespread occurrence of these diseases, which is estimated at over 300 million people worldwide, medical image analysis methods for~\ac{OCT} imaging have gained popularity in recent years. The automatic segmentation of retinal layer structures is of particular interest as it allows for the quantification, characterization and monitoring of retinal disorders over time. This remains a challenging task, as retinal layers can be heavily distorted in the presence of pathological biomarkers. In this context, the present paper focuses on providing more accurate retinal layer segmentations in pathological eyes, at clinically-relevant speeds.

A number of relevant methods on this topic can be found in the literature. Mayer et al.~\cite{Mayer2010} propose the use of a series of edge filters and denoising steps to extract layers in OCT cross-sections. In~\cite{Dufour2013}, a \ac{MRF}-based optimization with soft constraints is proposed to segment 7 retinal layers using volumetric information. Chen et al.~\cite{Chen2012} use a constrained graph-cut approach to segment layers and quantify fluid pockets in pathological OCTs. Overall, most of these methods face difficulties in segmenting all retinal layers accurately for subjects with pathological eyes.

% Our contribution
To this end, we present a novel strategy to overcome the above limitations and provide accurate results in a wider range of cases. Inspired by recent \ac{CNN} approaches for semantic segmentation~\cite{Ronneberger2015} and image classification~\cite{He2015}, we introduce a novel \ac{CNN} architecture that learns to segment retinal layers as a supervised regression problem. Our proposed network combines residual building blocks with dilated convolutions into an asymmetric U-shape configuration, and can segment multiple layers of highly pathological eyes in one shot. Using lower computational resources, our strategy achieves superior segmentation performance compared to both state-of-the-art deep learning architectures and other \ac{OCT} segmentation methods.

\section{Methods}
Our goal is to segment retinal cell layers in \ac{OCT} images. The main challenge in this task stems primarily from the highly variable and irregular shape of pathological eyes, and secondarily from the variable image quality (\ie signal strength and speckle noise) of clinical \ac{OCT} scans. Due to the image acquisition process, wherein each cross-section, or {\it Bscan} is acquired separately without a guaranteed global alignment, we opt to segment retinal layers at the Bscan level. This avoids the need for computationally intensive 3-dimensional convolutions~\cite{Cicek2016} and volumetric pre-processing (\ie registration and alignment).

In our approach, we treat the task of segmenting retinal layers as a regression problem. Given a Bscan image, $\mathcal{I}$, we wish to find a function $T:\mathcal{I} \to \mathcal{L}$, that maps each pixel in $\mathcal{I}$ to a label $\mathcal{L} \in \{0,1,2,3,4,5,6\}$ corresponding to an anatomical retinal cell layer region. As in~\cite{Dufour2013}, we consider the following six retinal layers: (1)~\ac{ILM} to \ac{NFL}, (2)~\ac{NFL} to \ac{GCL}, (3) \ac{GCL} and \ac{IPL}, (4)~\ac{INL} and \ac{OPL}, (5)~\ac{OPL} to \ac{IS/OS} Junction and (6)~\ac{IS/OS} Junction to \ac{BM}.

\subsection{Branch Residual U-Network}
Fully convolutional U-net style networks have established themselves as the state-of-the-art for binary segmentation and have been successfully used in a variety of biomedical applications~\cite{Ronneberger2015}. In such architectures, input images are convolved and downsampled level by level with exponentially increasing numbers of filters up to a predefined depth (\textit{descending branch}), from which they are subsequently upsampled and convolved to the original size (\textit{ascending branch}). Skip connections from corresponding levels transfer information from the descending to the ascending branch.

A number of important limits arise from this architecture. First, the largest possible object that can be segmented is defined by the cumulative receptive field of the network. According to our experiments, a regular U-net with \SI{3x3}{px} convolutions and a depth of 5 layers~\cite{Ronneberger2015}, will start exhibiting holes when segmenting objects with discontinuities wider than $3*2^5=\SI{96}{px}$. Second, due to the exponential growth of trainable parameters, the maximum depth of such a network is limited to 5-7 layers before the computational demands become intractable. Third, the convergence rate of a U-net tends to decrease as the network grows in depth. We attribute this to the vanishing gradient problem that affects deeper networks.

\begin{figure}[t!]
\centering
\includegraphics[width=0.99\textwidth]{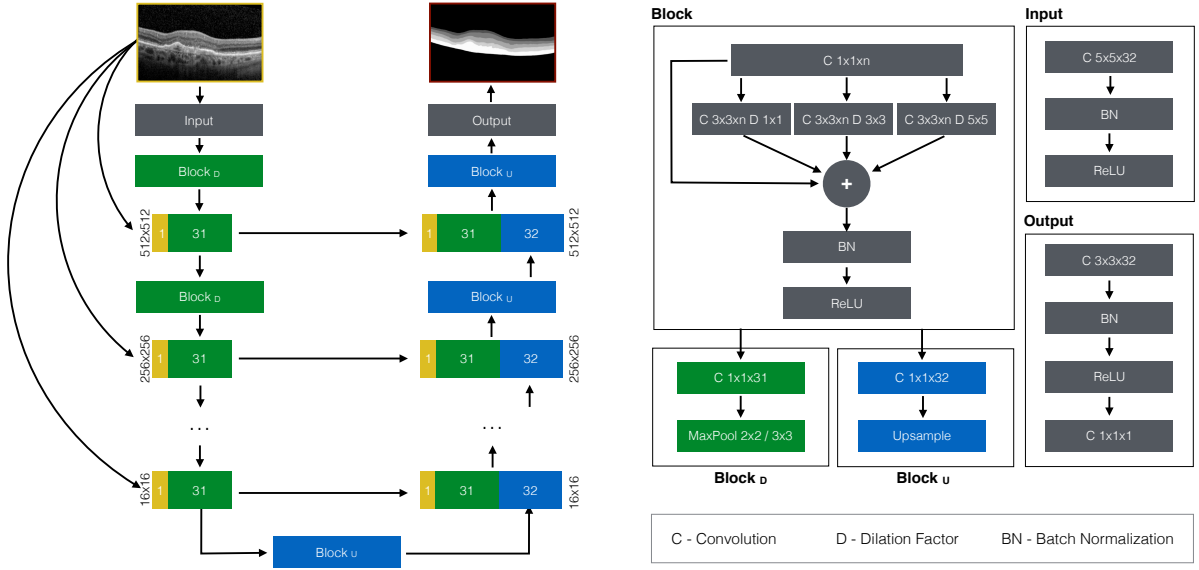}
\caption{
(left) Branch Residual U-Network (BRU-net). The descending branch takes a single Bscan as input and performs consecutive Block$_D$ operations. The ascending branch receives the output of the descending branch and performs consecutive Block$_U$ operations. The numbers indicate the number of filters output by each block. Skip connections connect each descending to each ascending level, while the original Bscan is provided to each level for context. The final output is a regressed layer class for each pixel of the input image.\\
(right) Block$_D$, Block$_U$, input and output blocks. The rectangles illustrate computations.}
\label{fig:network}
\end{figure}

We have designed our network to address each of these problems:
\begin{enumerate}
\item We use a building block based on dilated convolutions with dilation rates of $\{1, 3, 5\}$ to increase the effective receptive field of each network level and without increasing the number of trainable parameters. We enhance this block with residual connections~\cite{He2016,Wu2016} and batch normalization~\cite{Ioffe2015}, which are summed together with the dilated convolutions. Depending on the branch direction, each block ends with a max-pooling or upsampling operation. We denote those blocks as {\it Block$_D$} and {\it Block$_U$}, respectively.
\item We insert bottleneck connections between blocks to control the number of trainable parameters~\cite{Szegedy2015,Szegedy2016,Huang2016}. Furthermore, we increase the number of filters based on a capped Fibonacci sequence. We chose this sequence after experimenting with zero, constant and quadratic growth, as a good trade-off between network capacity and segmentation performance.
\item Finally, we add connections from the input image, downscaled to the appropriate size, to all levels in the ascending and descending branches.
\end{enumerate}
Combined, these result in a significant increase in the learning rate, segmentation accuracy and, due to the reduced number of parameters, processing speed. We name the resulting architecture {\it Branch Residual U-shape Network} (BRU-net). The precise architecture and building blocks are illustrated in Fig.~\ref{fig:network}.

Throughout our network, we employ $3\times 3$ convolutional kernels with $n$ filters where $n$ increases according to the Fibonacci sequence $\{32,64,96,160,256,416\}$, capped to a maximum of $416$ parameters per level. This avoids the larger growth of parameters encountered in traditional U-networks and allows for deeper networks. More specifically, our network requires \SI{21}{million} parameters for a depth of 5 levels and grows to \SI{55}{million} parameters for a depth of 6 levels. The corresponding U-net requires \SI{44}{million} and \SI{176}{million} parameters for the same depths, an increase of 2$\times$ and 3$\times$, respectively.

\subsection{Training}
\label{sec:train}
The block layout has been optimized using an evolutionary grid search strategy, by training two variants in parallel and selecting the best performer. To keep training time reasonable, the grid search is performed on a 4$\times$ subsampled dataset. This process is repeated 50 times, each one taking up to 30 minutes.

To increase convergence rate and reduce training time, we pre-initialize our network by training it as an autoencoder for 10 epochs, using a small set of 50 OCT volumes of healthy eyes, acquired from the same OCT device. This set is distinct from the volumes we use for segmentation. To avoid learning the identity function, we disable the skip connections of the network during this process.

The output of the network is an image with the same size as the input Bscan. Each pixel of the output image is assigned a value between 0 and 6 which corresponds to the identity of its corresponding retinal layer. We train the network to minimize the pixel-wise \ac{MSE} loss between the predicted segmentation and the ground truth. This loss penalizes anatomically implausible segmentations (e.g. class 6 next to 0) more than plausible ones (e.g. class 1 next to 0). We rely this asymmetry to ensure segmentation continuity.\ie The network parameters are updated via back-propagation and the Adam optimization process with the infinity norm.~\cite{Kingma2014}.

Each fold is trained for a maximum of 150 epochs. We start training with an initial learning rate of $10^{-3}$ and reduce it by a factor of 2 if the \ac{MSE} loss does not improve for 5 consecutive epochs, down to a minimum of $10^{-7}$. We interrupt the training early if the \ac{MSE} loss stops improving for 25 consecutive epochs. Using a dedicated validation set, comprising 10\% of the training set, we evaluate the \ac{MSE} loss to adaptively set the learning rate and perform early stopping. At the end of the training procedure, we use the network weights of the epoch with the lowest validation loss to evaluate images in the test set.

\section{Experimental Results}
\label{sec:exp}
\begin{figure}[t!]
\centering
\includegraphics[width=0.30\textwidth]{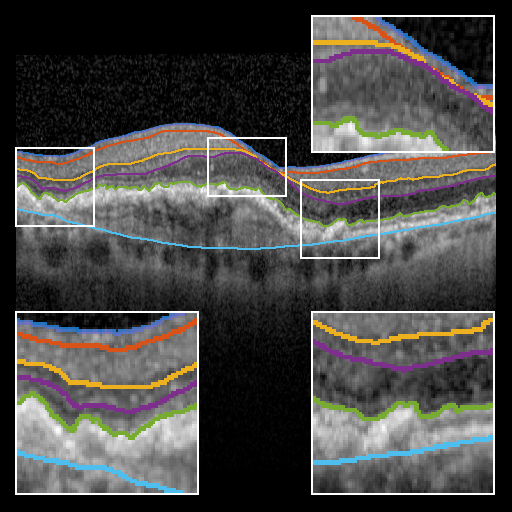}
\includegraphics[width=0.30\textwidth]{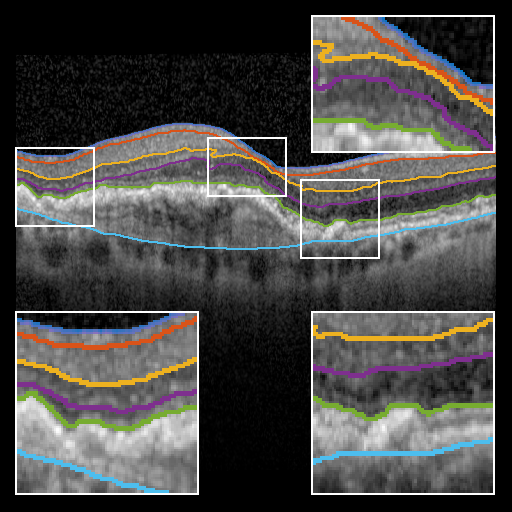}
\includegraphics[width=0.30\textwidth]{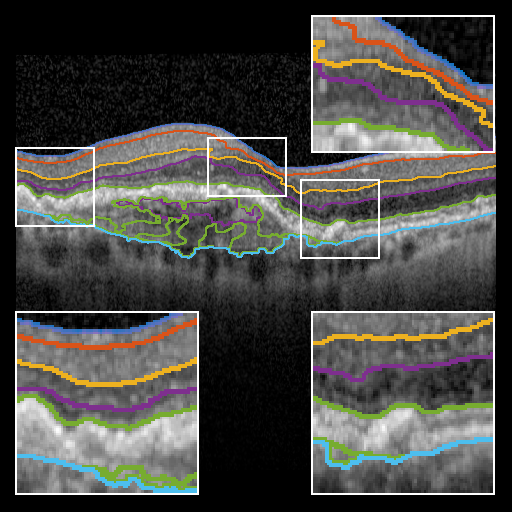}

\includegraphics[width=0.30\textwidth]{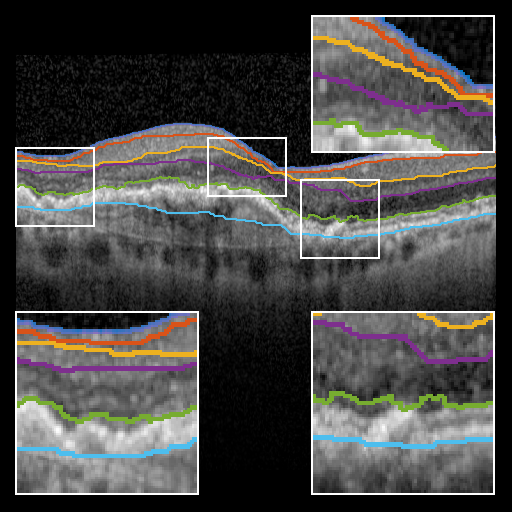}
\includegraphics[width=0.30\textwidth]{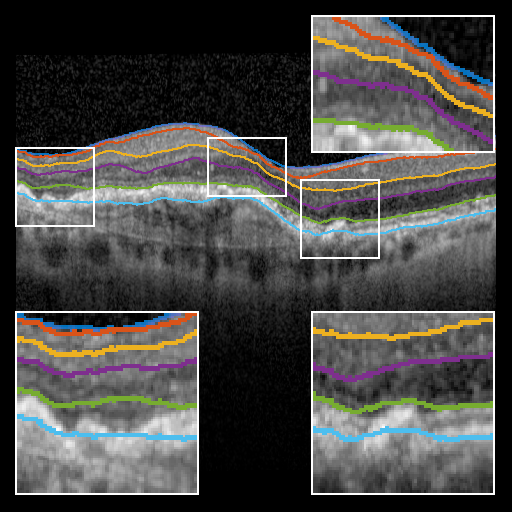}
\includegraphics[width=0.30\textwidth]{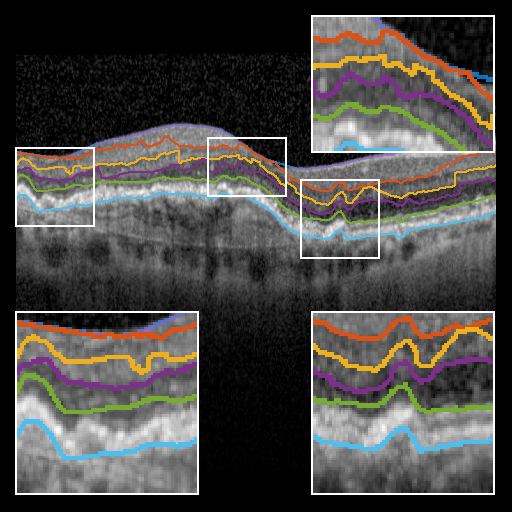}
\caption{Qualitative comparison of each segmentation approach. Top row, left to right: ground truth, BRU-net, U-net; bottom row, left-to-right: Dufour et al., Chen et al., Mayer et al. Only BRU-net is able to segment the \ac{BM} layer under the pathological region. The smaller receptive field of U-net results in discontinuities. Further qualitative results are provided in the supplementary material.}
\label{fig:qualitative}
\end{figure}

A trained ophthalmologist collected 20 macular OCT volumes from pathological subjects using a Heidelberg Spectralis OCT device (Heidelberg Engineering AG, Heidelberg, Germany). Each volume comprises 49, $512 \times 496$ Bscans, with a lateral (x-y) resolution of \SI{15}{\micro \meter} and an axial (z-) resolution of \SI{3.9}{\micro \meter}. No volumes or Bscans were removed from our initial acquisition, to maintain the complete range of image quality observed in the clinic. For each Bscan in each volume, manually segmented ground truth layers were provided by the ophthalmologist.

We split our dataset into 5 equally sized subsets, each using 16 patients for training and 4 for testing. We repeat this process for each of those subsets for a 5-fold cross-validation. In each fold, the training set contains 784 training samples (Bscans), which we double to 1568 by flipping horizontally, taking advantage of the bilateral symmetry of the eye. The Bscans are first padded with a black border to a size of 512$\times$512 pixels and then augmented with affine transformations, additive noise, Gaussian blur and gamma adjustments. Training is performed on batches of 8 Bscans at a time. Finally, the output image is quantized to integer values (0 to 6) without further post-processing.
\begin{figure}[t!]
\centering
\includegraphics[width=0.49\textwidth,height=150pt]{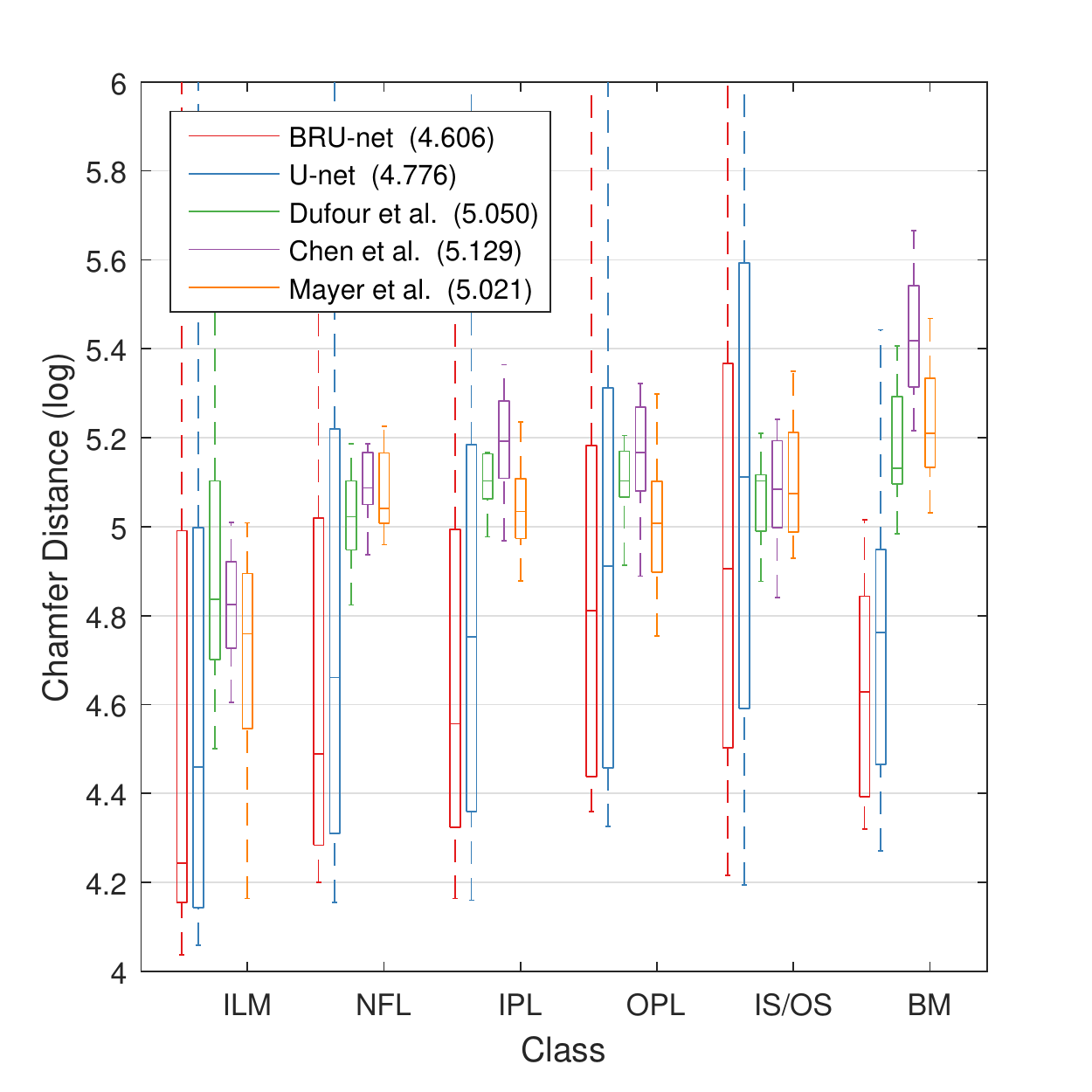}
\includegraphics[width=0.49\textwidth,height=150pt]{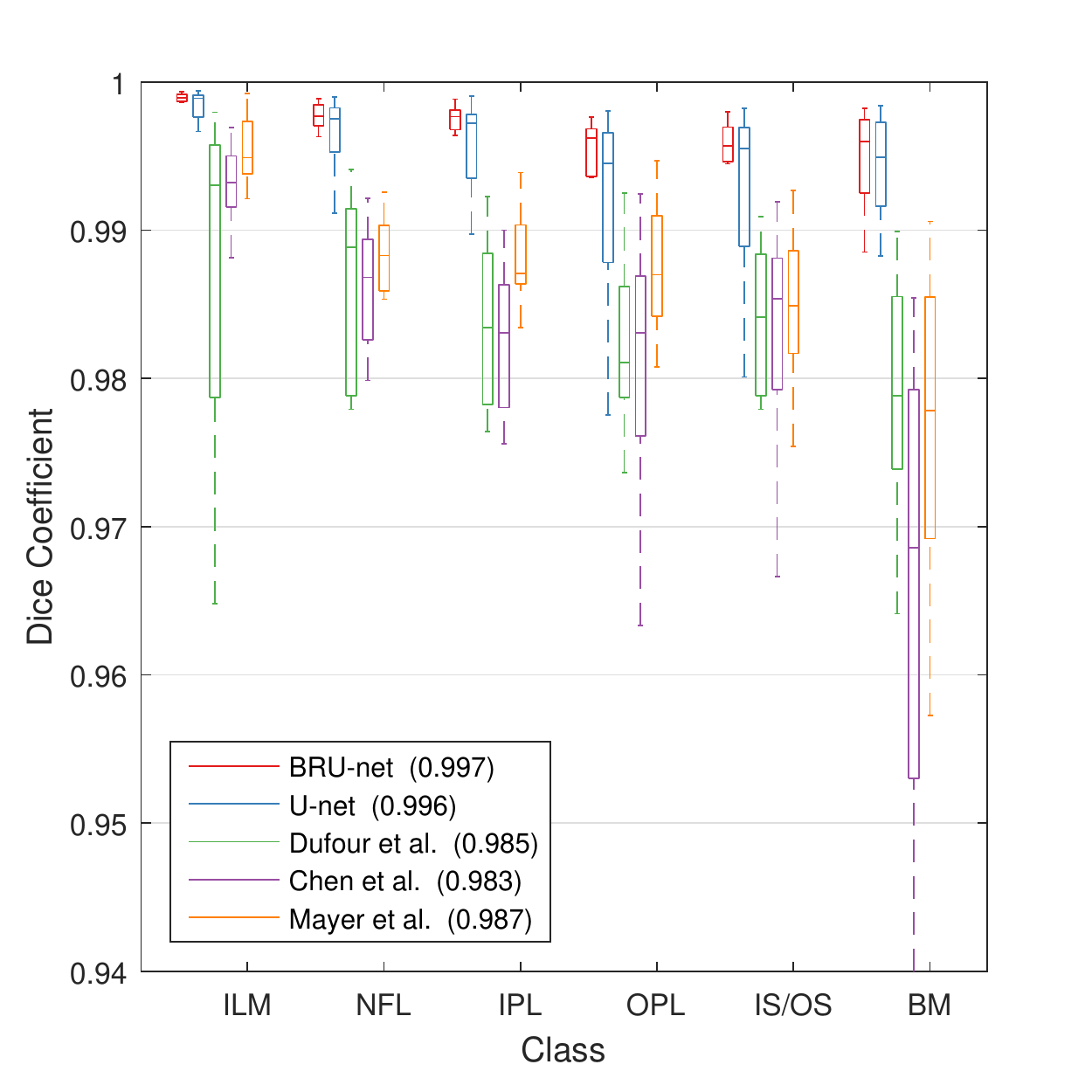}
\caption{Quantitative comparison of segmentation accuracy per layer, using (left) the Chamfer distance error and (right) the Dice score.}
\label{fig:quantitative}
\end{figure}

To evaluate BRU-net, we compare it with the 3D methods of Dufour~\cite{Dufour2013}, Chen et. al.~\cite{Chen2012}, and the 2D method of Mayer et. al.~\cite{Mayer2010} on the same dataset. Additionally, we train a traditional U-net configuration~\cite{Ronneberger2015} using the procedure described above. Fig.~\ref{fig:qualitative} provides a qualitative comparison of the results.

To quantify those results, we make use of two metrics: (1) the Chamfer distance~\cite{Butt98} between each ground truth layer boundary and the boundary produced by a given method and (2) the Dice score of each predicted layer surface. Note that BRU-net is not constrained to convex shapes. Since pathological retinal layers may be non-convex, other metrics that rely on pixel distances are ill-suited for this problem. Fig.~\ref{fig:quantitative} demonstrates the performance of each of the evaluated methods.

\begin{figure}[t!]
\centering
\includegraphics[width=0.49\textwidth,height=120pt]{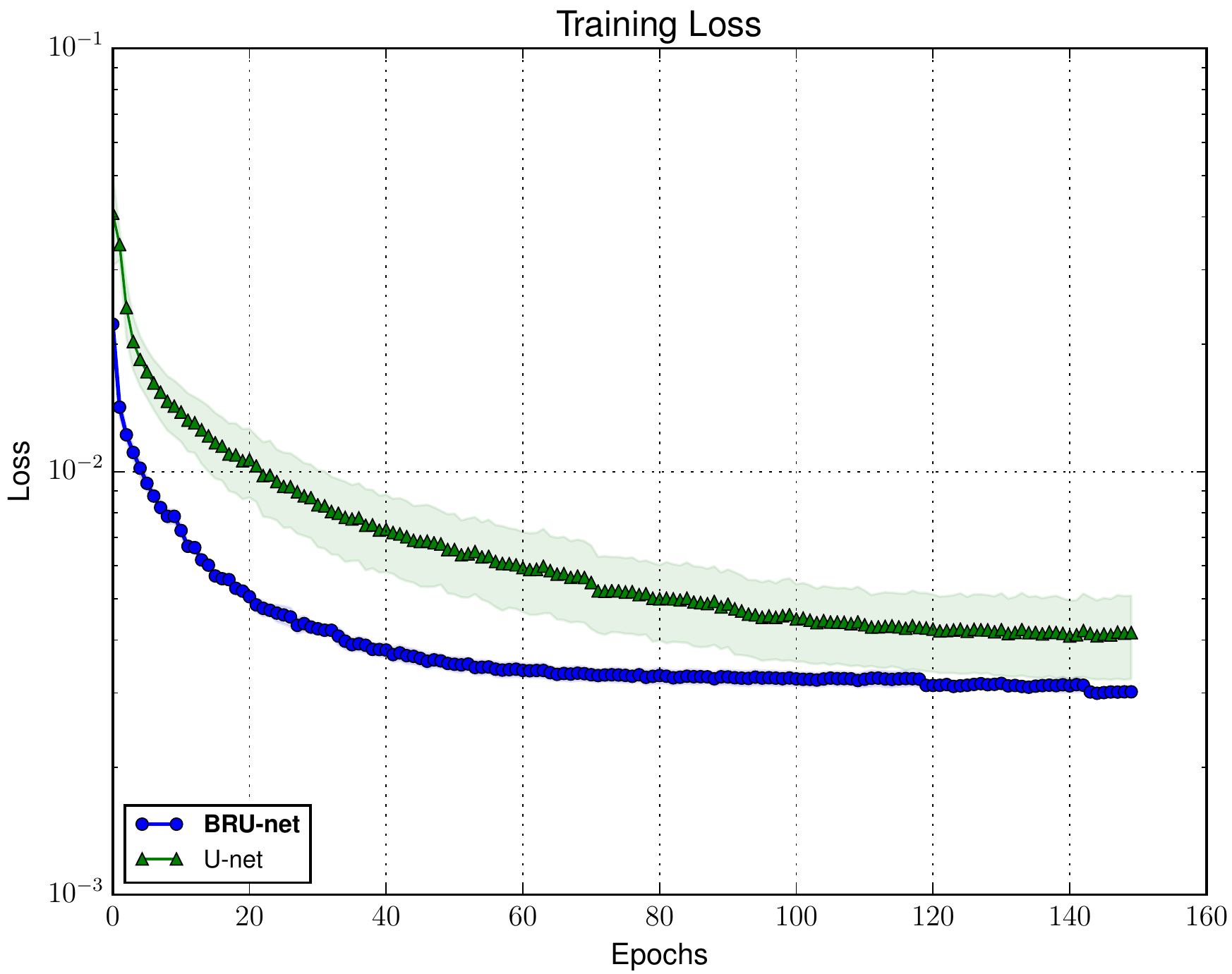}
\includegraphics[width=0.49\textwidth,height=120pt]{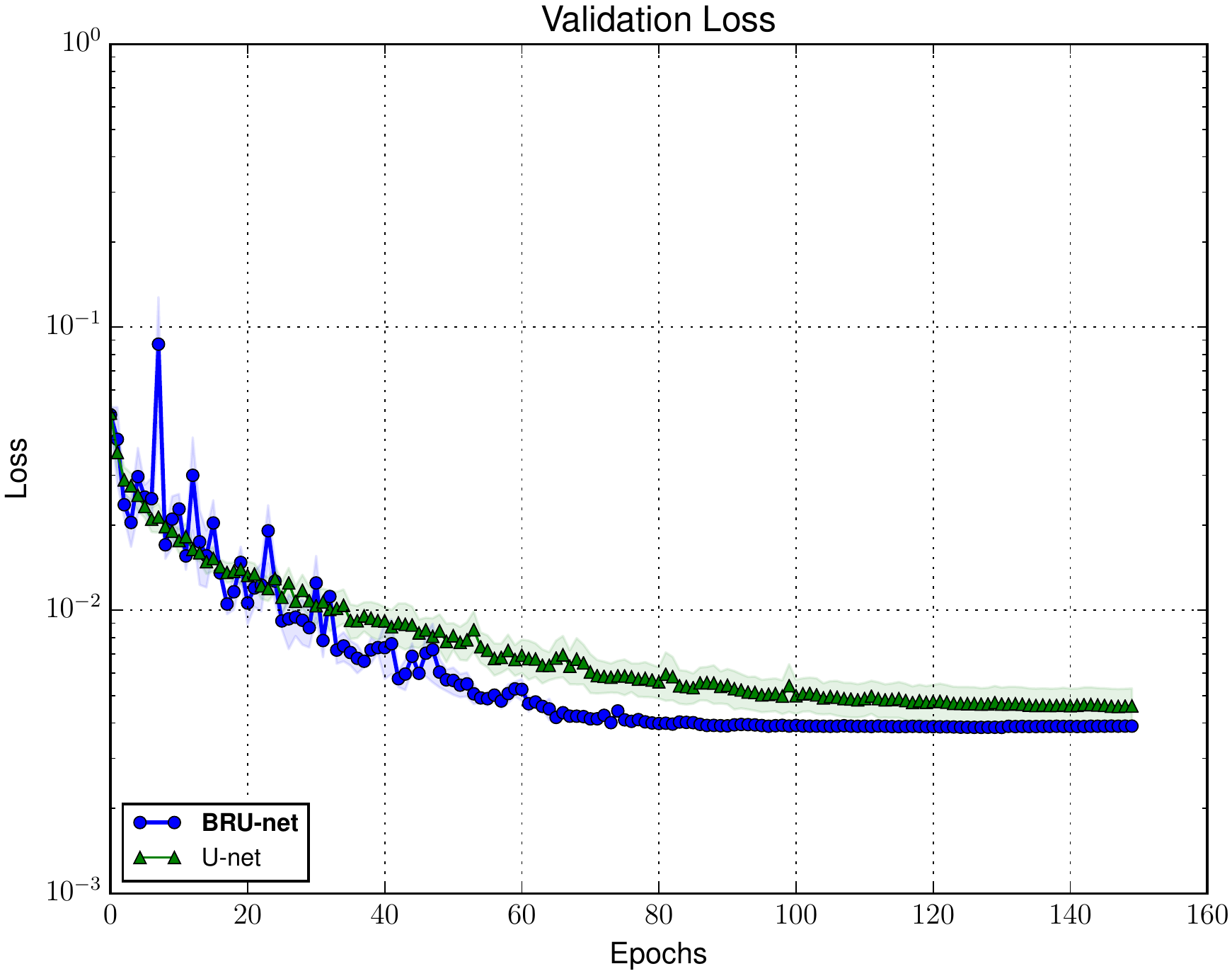}
\caption{Training loss (left) and validation loss (right) comparison between BRU-net and U-net. BRU-net exhibits faster convergence speed and lower loss compared to U-net.}
\label{fig:training}
\end{figure}

Fig.~\ref{fig:training} displays the mean training and validation loss of the 5-folds over time for both BRU-net and U-net. In both the training and validation sets, BRU-net achieves faster convergence and slightly better \ac{MSE} loss.

We evaluated the statistical significance of those results using paired t-tests between BRU-net and each baseline. The resulting p-values indicate statistically significant results between BRU-net and every other baseline except U-net:

\begin{center}
\begin{tabular}{p{2.5cm} | p{2cm}  p{2.2cm}  p{2.2cm}  p{2.2cm}}
~~            & ~~U-net    & Dufour et. al. & Chen et. al. & Mayer et. al. \\
\hline
\\[-1em]
~~p (Dice)    & ~~1.05e-01 & 2.03e-04       & 3.19e-05     & 1.08e-05 \\
~~p (Chamfer) & ~~1.09e-01 & 2.70e-04       & 3.18e-03     & 9.49e-03 \\ 
\end{tabular}
\end{center}

Finally, we estimated the total runtime for each method. To process a single volume, BRU-net requires \SI{5}{s} (Python), compared to \SI{7}{s} for U-net (Python), \SI{85}{s} for Mayer et al. (Matlab), \SI{150}{s} for Dufour et al. (C++), and  and \SI{216}{s} for Chen et al (C++). The results were calculated on the same system using a \SI{3.9}{GHz} Intel 6600K processor and a Nvidia 1080GTX GPU.

\section{Conclusion}
We have presented a method for performing layer segmentation on \ac{OCT} scans of highly pathological retinas. Inspired by recent advances in computer vision, we have designed a novel fully-convolutional \ac{CNN} architecture that can segment multiple layers in one shot. We have compared our method to several baselines and demonstrated qualitative and quantitative improvements in both segmentation accuracy and computational time on a dataset of late-stage \ac{AMD} patients. Given the robustness of this approach on pathological cases, we plan to investigate how retinal layers change over time in the presence of specific diseases. \comment{Crucially, provided suitable ground truth data, our approach can be applied to other types of diseases, tissue or imaging modalities. We believe similar approaches to have significant clinical potential for the quantification of disease biomarkers, and intend to pursue this research avenue in the near future.}
\bibliographystyle{plain}
\bibliography{miccai_ref}

\end{document}